# Structured Convex Optimization under Submodular Constraints

**Kiyohito Nagano**
Dept. of Complex and Intelligent Systems
Future University Hakodate
k_nagano@fun.ac.jp

**Yoshinobu Kawahara**
The Institute of Scientific and Industrial Research
Osaka University
kawahara@ar.sanken.osaka-u.ac.jp

## Abstract

A number of discrete and continuous optimization problems in machine learning are related to convex minimization problems under submodular constraints. In this paper, we deal with a submodular function with a directed graph structure, and we show that a wide range of convex optimization problems under submodular constraints can be solved much more efficiently than general submodular optimization methods by a reduction to a maximum flow problem. Furthermore, we give some applications, including sparse optimization methods, in which the proposed methods are effective. Additionally, we evaluate the performance of the proposed method through computational experiments.

## 1 Introduction

A submodular function is a fundamental tool in discrete optimization, machine learning and other related fields and has been recognized as an interesting subject of research. A submodular function is known to be a discrete counterpart of a convex function (Lovász [17]). Especially, the *submodular function minimization* problem is an elemental problem, and many combinatorial problems arising in machine learning, such as clustering [25, 24], image segmentation [31] and feature selection [2], can be reduced to this problem.

For example, Narasimhan, Joic and Bilmes [25] showed that clustering problems with some specific natural criteria, such as the minimum description length, can be solved as the problem of minimizing a symmetric submodular function. Also, Bach [2] recently showed that many of the known structured-sparsity inducing norms can be interpreted as continuous relaxations, called the Lovász extensions, of submodular functions. Based on this correspondence relationship, proximal operators, which are required for learning with structured regularization, can be computed as minimum-norm-point problems on submodular polyhedra.

Similarly to convex functions, submodular functions can be exactly minimized in polynomial time. The fastest known algorithm of Orlin [27] runs in $O(n^5\text{EO}+n^6)$ time, where $n$ is the size of the ground set and EO is the time for function evaluation. On the other hand, the minimum norm point algorithm (Fujishige [9]) is usually much faster in practice [10], although it has worse time complexity. However, the existing algorithms for the general submodular minimization problem, even including the minimum norm point algorithm, do not scale sufficiently to large problems from a practical point of view.

Meanwhile, it is known that submodular function minimization problems can be solved more efficiently when the submodular functions have particular structure. For symmetric submodular functions, Queyranne [29] gave a minimization algorithm that runs in $O(n^3\text{EO})$. Also recently, Stobbe and Krause [31] introduced a decomposable submodular function and developed the Smoothed Lovász Gradient (SLG) algorithm, which is based on the smoothing technique of Nesterov [26] and the discrete convexity of a submodular function. In addition, Jegelka et al. [14] introduced a generalized graph cut function, which generalizes a large subfamily of submodular functions, and proposed an efficient network flow based minimization algorithm.

In this paper, we consider a separable convex optimization problem over a base polyhedron, which is a discrete structure determined by a submodular function. Separable convex optimization under submodular constraints is related to various discrete and continuous optimization problems, including network analysis methods [23], sparse learning methods [2], and approximation algorithms for NP-hard combinatorial optimization problems [13]. For a general submodular function, separable convex optimization problems can be solved within the same running time as submodular function minimization [6, 21], that is, $O(n^5\text{EO} + n^6)$ time. Thus, such algorithms are impractical when the size of the ground set is large. Even though the minimum norm point algorithm [9] and its weighted version [22] would solve such quadratic minimization problem much faster, it does not have good time complexity bounds and still does not scale to large problems.

We show that if a submodular function has a specific graph structure, the convex optimization problem can be solved efficiently with the aid of a general framework of the decomposition algorithm [9, 22] and network flow algorithms [11, 12, 28]. We develop a parametrized directed graph structure that determines a parametric submodular function minimization problem, and show that the decomposition algorithm can be performed successfully by computing the maximal minimum cuts iteratively. Furthermore, we mention that several machine learning applications can be solved in this convex optimization problem. We remark that the proposed method can deal with a relatively general submodular function and various separable convex objective functions.

The remainder of the paper is organized as follows. In Section 2, we provide the definitions of basic concepts and give a definition of a convex optimization problem under submodular constraints. In Section 3, we give examples of submodular functions that have good graph structures. In Section 4, we show some optimization problems related to separable convex optimization problems under submodular constraints. In Section 5, we describe a general decomposition algorithm for solving separable convex optimization problems, and in Section 6, we further show that structured convex optimization problems under submodular constraints can be solved efficiently with the aid of network flow algorithms. Finally, we show some empirical results of computational experiments in Section 7, and give concluding remarks in Section 8.

## 2 Submodular functions and convex optimization problems

We give basic definitions of a submodular function and related concepts (for details on the theory of submodular functions, see [9, 30]). Then, we give the definition of a convex optimization problem under submodular constraints.

### 2.1 Submodular functions and related polyhedra

Let $V = \{1, \ldots, n\}$ be a given set of $n$ elements, and let $g : 2^V \to \mathbb{R}$ be a real-valued function defined on all the subset of $V$. Such a function $g$ is called a *set function* with a ground set $V$. The set function $g : 2^V \to \mathbb{R}$ is called *submodular* if

$$g(S) + g(T) \geq g(S \cup T) + g(S \cap T), \forall S, T \subseteq V. \quad (1)$$

A set function $g$ is called *supermodular* if $-g$ is submodular. A set function is called *modular* if it always satisfies (1) with equality. A set function is called *nondecreasing* if $g(S) \leq g(T)$ for any $S, T \subseteq V$ with $S \subseteq T$. For an $n$-dimensional vector $\boldsymbol{a} \in \mathbb{R}^n$ with components $a_i$, $i \in V$, and a subset $S \subseteq V$, we denote $a(S) = \sum_{i \in S} a_i$. For convenience, we let $a(\varnothing) = 0$. A set function $a : 2^V \to \mathbb{R}$ corresponding to the vector $\boldsymbol{a}$ is a modular function.

**Submodular function minimization**

A submodular function minimization problem is a fundamental unifying discrete optimization problem. For a submodular function $g : 2^V \to \mathbb{R}$, the submodular function minimization problem asks for finding a subset $S \subseteq V$ that minimizes $f(S)$. This problem is known to be solvable in polynomial time, and the fastest known polynomial time algorithm [27] that runs in $O(n^5\text{EO} + n^6)$ time, where EO is the time of one function evaluation of $g$. The algorithms for general submodular function minimization are impractical when $n = |V|$ is large. In addition, the minimum norm point algorithm [9] is known to be usually much faster in practice, although it has worse time complexity.

Let $\operatorname{Arg\,min} g \subseteq 2^V$ denote the family of all minimizers of $g$. That is, $\operatorname{Arg\,min} g = \{S^* \subseteq V : f(S^*) = \min_S f(S)\}$. For $S^*, T^* \in \operatorname{Arg\,min} g$, the submodularity of $g$ implies that $S^* \cup T^*, S^* \cap T^* \in \operatorname{Arg\,min} g$. Thus, there exist the (unique) minimal minimizer and the (unique) maximal minimizer of $g$. Many submodular function minimization algorithms can be modified to find the maximal minimizer and/or the minimal minimizer (see, e.g., [21]).

**Base polyhedron**

For a submodular function $g : 2^V \to \mathbb{R}$ with $g(\varnothing) = 0$, the *submodular polyhedron* $\mathrm{P}(g) \subseteq \mathbb{R}^n$ and the *base polyhedron* $\mathrm{B}(g) \subseteq \mathbb{R}^n$ are given by

$$\mathrm{P}(g) = \{\boldsymbol{x} \in \mathbb{R}^n : x(S) \leq g(S) \ (\forall S \subseteq V)\},$$
$$\mathrm{B}(g) = \{\boldsymbol{x} \in \mathrm{P}(g) : x(V) = g(V)\}.$$

Figure 1 illustrates examples of the base polyhedra. $\mathrm{B}(g)$ is determined by $2^n - 2$ inequalities and one equality. We see that $\mathrm{B}(g)$ is nonempty and bounded. The base polyhedron $\mathrm{B}(g)$ is included in the nonnegative orthant $\mathbb{R}^n_{\geq 0}$ if and only if $g$ is nondecreasing.

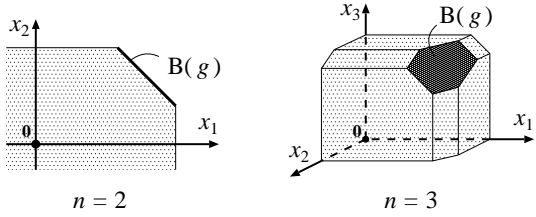

Figure 1: Examples of base polyhedra

## 2.2 Convex optimization under submodular constraints

Throughout this paper, we suppose that set function $f : 2^V \to \mathbb{R}$ is submodular and satisfies $f(\varnothing) = 0$. Let $w_i : \mathbb{R} \to \mathbb{R}$ be a convex function for each $i \in V$. We consider the separable convex function minimization problem over the base polyhedron:

$$\min_{\boldsymbol{x} \in \mathrm{B}(f)} \sum_{i \in V} w_i(x_i). \qquad (2)$$

It is known that a number of optimization problems of this form are equivalent.

**Theorem 1** (Nagano and Aihara [22])**.** *Suppose that $f : 2^V \to \mathbb{R}$ is a nondecreasing submodular function with $f(\varnothing) = 0$. Let $\boldsymbol{b} \in \mathbb{R}^n$ be a positive vector, and let $w_0 : \mathbb{R} \to \mathbb{R}$ be a differentiable and strictly convex function. The following problems* (1.a) – (1.f) *have the same (optimal) solution:*

*problem* (1.a) $\quad \min_{\boldsymbol{x} \in \mathrm{B}(f)} \sum_{i=1}^{n} \frac{x_i^2}{b_i}$;

*problem* (1.b) $\quad \min_{\boldsymbol{x} \in \mathrm{B}(f)} \sum_{i=1}^{n} \frac{x_i^{p+1}}{b_i^p}$ *for* $p > 0$;

*problem* (1.c) $\quad \max_{\boldsymbol{x} \in \mathrm{B}(f)} \sum_{i=1}^{n} \frac{x_i^{p+1}}{b_i^p}$ *for* $p < 0$ *with* $p \ne -1$;

*problem* (1.d) $\quad \max_{\boldsymbol{x} \in \mathrm{B}(f)} \sum_{i=1}^{n} b_i \ln x_i$;

*problem* (1.e) $\quad \min_{\boldsymbol{x} \in \mathrm{B}(f)} \sum_{i=1}^{n} (x_i \ln \frac{x_i}{b_i} + b_i - x_i)$;

*problem* (1.f) $\quad \min_{\boldsymbol{x} \in \mathrm{B}(f)} \sum_{i=1}^{n} x_i g_0(\frac{b_i}{x_i})$.

In view of Theorem 1, we focus on the case where the objective function is quadratic. For a positive vector $\boldsymbol{b} \in \mathbb{R}^n$, we mainly deal with problem (1.a). By using the following two observations, w.l.o.g., we can assume that the submodular function $f$ is nondecreasing.

**Lemma 2.** *For any $\beta \in \mathbb{R}$, $\boldsymbol{x}^*$ is optimal for $\min\{\sum_i \frac{x_i^2}{b_i} : x \in \mathrm{B}(f)\}$ if and only if $\boldsymbol{x}^* + \beta \boldsymbol{b}$ is optimal for $\min\{\sum_i \frac{x_i^2}{b_i} : x \in \mathrm{B}(f + \beta b)\}$.*

**Lemma 3.** *Set $\beta := \max\{0, \max_{i=1,\dots,n} \frac{f(V \setminus \{i\}) - f(V)}{b_i}\}$. Then $f + \beta b$ is a nondecreasing submodular function.*

Problem (1.a) is known as the lexicographically optimal base problem [8]. If $\boldsymbol{b}$ is the all-one vector, problem (1.a) becomes the minimum norm base problem. For a general submodular function, problem (1.a) can be solved within the same running time as the submodular function minimization [6, 21], that is, $\mathrm{O}(n^5\mathrm{EO} + n^6)$ time, where EO is the time of one function evaluation. Thus, such algorithms are impractical when $n = |V|$ is large. Although the minimum norm point algorithm [9] and its weighted version [22] can solve problem (1.a) much faster, it has worse time complexity and still does not scale to large problems.

In this paper, we point out that if the function $f$ has a good graph structure, problem (1.a) can be solved efficiently with the aid of network flow algorithms. Furthermore, we show a number of applications of the convex optimization problem (1.a).

## 3 Structured submodular functions and minimization problems

Many basic submodular functions can be represented by using graphs. In such cases, a minimum cut algorithm, which runs much faster in practice, is useful to solve submodular optimization problems.

In this section, we will see some examples of submodular functions with directed graph structures, which are important from the viewpoint of applications.

### 3.1 Minimizing graph cut functions

In this subsection, we will see that an *s-t* cut function $\kappa_{s\text{-}t}$ and a generalized graph cut function $\gamma$ of [14], both of which are submodular, can be minimized efficiently with the aid of network flow algorithms. In particular, we will see that the maximal minimizer can be computed efficiently in both cases. In the general algorithm described in Section 5, the maximal minimizer of a submodular function has to be computed.

**Minimum cut problem**

We start with the minimum *s-t* cut problem. Let $\mathcal{G} = (\{s\} \cup \{t\} \cup \mathcal{V}, \mathcal{E})$ be a directed graph, where $s$ is a special source node, $t$ is a special sink node, $\mathcal{V}$ is a set of other nodes, and $\mathcal{E}$ is a set of directed edges. For each $e \in \mathcal{E}$, a nonnegative capacity value $c(e)$ is assigned. An *s-t* cut is an ordered bipartition $(\mathcal{V}_1, \mathcal{V}_2)$ of the node set of $\mathcal{G}$ such that $s \in \mathcal{V}_1$ and $t \in \mathcal{V}_2$. Clearly, any *s-t* cut can be expressed as $(\{s\} \cup S, \{t\} \cup (\mathcal{V} \setminus S))$ for some $S \subseteq \mathcal{V}$. For an *s-t* cut $(\{s\} \cup S, \{t\} \cup (\mathcal{V} \setminus S))$, its capacity $\kappa_{s\text{-}t}$ is defined by

$$\kappa_{s\text{-}t}(S) = \sum\{c(e) : e \in \delta_{\mathcal{G}}^{\mathrm{out}}(\{s\} \cup S)\} \qquad (3)$$

for each $S \subseteq \mathcal{V}$, where $\delta_{\mathcal{G}}^{\text{out}}(\mathcal{V}')$ is a set of edges leaving $\mathcal{V}' \subseteq \mathcal{V}$ in $\mathcal{G}$. The minimum cut problem asks for finding an $s$-$t$ cut of $\mathcal{G}$ that minimizes the capacity. The set function $\kappa_{s\text{-}t} : 2^{\mathcal{V}} \to \mathbb{R}$, which is called an *s-t cut function*, is known to be submodular. Therefore, the minimum cut problem is a special case of a submodular function minimization problem.

The minimum cut problem is closely related to the maximum flow problem, which is a fundamental problem in combinatorial optimization [1]. It can be solved quite efficiently. For example, it can be solved in $O(|\mathcal{V}||\mathcal{E}|\log(|\mathcal{V}|^2/|\mathcal{E}|))$ time [12] or $O(|\mathcal{V}||\mathcal{E}|)$ time [28].

As $\kappa_{s\text{-}t}$ is submodular, there exists the maximal minimizer $S^*_{\max}$ of $\kappa_{s\text{-}t}$. The $s$-$t$ cut $(\{s\} \cup S^*_{\max}, \{t\} \cup (\mathcal{V} \setminus S^*_{\max}))$ is called the *maximal minimum s-t cut*. Once a maximal flow is computed, we can obtain the maximal minimum $s$-$t$ cut in additional $O(|\mathcal{V}| + |\mathcal{E}|)$ time (we just need to consider the set of nodes reachable to the sink $t$ and its complement in the residual network [1]). The minimal minimum $s$-$t$ cut can be defined and computed in a similar way.

**Lemma 4.** *The maximal minimizer of the s-t cut function $\kappa_{s\text{-}t} : 2^{\mathcal{V}} \to \mathbb{R}$ defined in (3) can be computed in $O(|\mathcal{V}||\mathcal{E}|\log(|\mathcal{V}|^2/|\mathcal{E}|))$ time, or, $O(|\mathcal{V}||\mathcal{E}|)$ time.*

**Generalized graph cut functions**

Next we give a definition of the generalized graph cut function $\gamma : 2^V \to \mathbb{R}$ of Jegelka et al. [14], which generalizes a large subfamily of submodular functions.

Let $\mathcal{G} = (\{s\} \cup \{t\} \cup \mathcal{V}, \mathcal{E})$ be a directed graph with nonnegative edge capacities $c(e) \geq 0$ ($e \in \mathcal{E}$). Suppose that the set $\mathcal{V}$ is partitioned as $\mathcal{V} = V \cup U$, where $V = \{1, \ldots, n\}$ is a set of nodes, each of which may become a source, and $U$ is a set of auxiliary nodes ($U$ can be empty). Figure 2 illustrates an example of the graph $\mathcal{G} = (\{s\} \cup \{t\} \cup V \cup U, \mathcal{E})$. A generalized graph cut function [14] $\gamma : 2^V \to \mathbb{R}$ is defined by

$$\gamma(S) = \min_{W \subseteq U} \sum \{c(e) : e \in \delta_{\mathcal{G}}^{\text{out}}(\{s\} \cup S \cup W)\} \quad (4)$$

for each $S \subseteq V$. If $U$ is empty, the function $\gamma$ coincides with the function $\kappa_{s\text{-}t}$ defined in (3). The submodularity of $\gamma$ can be derived from the classical result of Megiddo [20] on network flow problems with multiple terminals (for details, see the appendix of this paper).

Let us consider the minimization of $\gamma : 2^V \to \mathbb{R}$. By the definition of $\gamma$, the value $\gamma^* := \min_{S \subseteq V} \gamma(S)$ is equal to the capacity of a minimum $s$-$t$ cut in $\mathcal{G}$. For any minimum $s$-$t$ cut $(\{s\} \cup \mathcal{P}, \{t\} \cup (V \cup U \setminus \mathcal{P}))$ in $\mathcal{G}$, we have $\gamma(\mathcal{P} \cap V) = \gamma^*$ and thus $\mathcal{P} \cap V$ is a minimizer of $\gamma : 2^V \to \mathbb{R}$. Therefore, a minimizer of $\gamma$ can be computed by solving the minimum $s$-$t$ cut problem on $\mathcal{G} = (\{s\} \cup \{t\} \cup V \cup U, \mathcal{E})$.

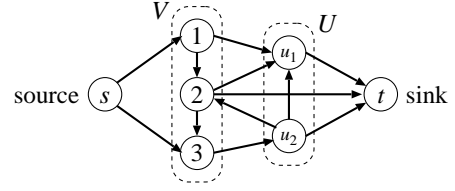

Figure 2: A directed graph $\mathcal{G} = (\{s\} \cup \{t\} \cup V \cup U, \mathcal{E})$ that generates a generalized graph cut function $\gamma : 2^V \to \mathbb{R}$

Conversely, let $S^* \subseteq V$ be a minimizer of $\gamma$, and let $W^*$ be a subset $W \subseteq U$ that attains the minimum in the right hand side of (4) with respect to $S = S^*$. Then $S^* \cup W^* \subseteq \mathcal{V}$ minimizes the $s$-$t$ cut function (see [14]).

Therefore, given the maximal minimum $s$-$t$ cut $(\{s\} \cup \mathcal{P}^*_{\max}, \{t\} \cup (V \cup U \setminus \mathcal{P}^*_{\max}))$, the subset $\mathcal{P}^*_{\max} \cap V$ is the maximal minimizer of $\gamma$.

**Lemma 5.** *The maximal minimizer of the generalized graph cut function $\gamma : 2^V \to \mathbb{R}$ defined in (4) can be computed in $O(|\mathcal{V}||\mathcal{E}|\log(|\mathcal{V}|^2/|\mathcal{E}|))$ time, or, $O(|\mathcal{V}||\mathcal{E}|)$ time, where $\mathcal{V} = V \cup U$.*

### 3.2 Transformed graph cut functions

We define a transformed graph cut function, and we show that the function can be regarded as an $s$-$t$ cut function defined in Subsection 3.1. In Subsection 4.1, we will see that the convex minimization problem (1.a) under the constraints of this function is related to the densest subgraph problem.

Let $G = (V, E)$ be a directed graph with node set $V = \{1, \ldots, n\}$ and edge set $E$. Given nonnegative edge capacities $c(e)$ ($e \in E$), a cut function $\kappa : 2^V \to \mathbb{R}$ defined by $\kappa(S) = \sum \{c(e) : e \in \delta_G^{\text{out}}(S)\}$ for each $S \subseteq V$ is submodular. Let $\boldsymbol{a} \in \mathbb{R}^n$. Then, a *transformed graph cut function* $\kappa_a : 2^V \to \mathbb{R}$ defined by

$$\kappa_a = \kappa + a$$

is also submodular.

Let us see that the function $\kappa_a : 2^V \to \mathbb{R}$ can be regarded as an $s$-$t$ cut function on a new graph $\mathcal{G}_{\boldsymbol{a}}$.

Define $A_+ = \{i \in V : a_i > 0\}$ and $A_- = \{i \in V : a_i < 0\}$. By adding new nodes $s, t$ and new edges $E_+ \cup E_-$ to $G$, we construct a new directed graph $\mathcal{G}_{\boldsymbol{a}} = (\{s\} \cup \{t\} \cup V, E \cup E_+ \cup E_-)$, where $E_+ = \{(i, t) : i \in A_+\}$ and $E_- = \{(s, i) : i \in A_-\}$. The capacities of new edges are determined as follows: we set $c(i, t) = a_i$ ($\geq 0$) for each $(i, t) \in E_+$, and set $c(s, i) = -a_i$ ($\geq 0$) for each $(s, i) \in E_-$. Figure 3 shows the construction of $\mathcal{G}_{\boldsymbol{a}}$.

For an $s$-$t$ cut $(\{s\} \cup S, \{t\} \cup (V \setminus S))$ of $\mathcal{G}_{\boldsymbol{a}}$, its capacity

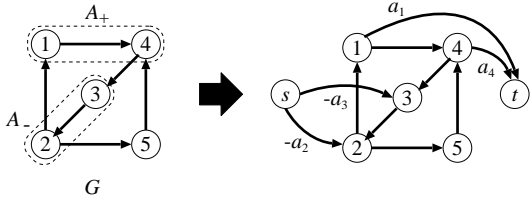

Figure 3: Construction of the directed graph $\mathcal{G}_{\boldsymbol{a}}$

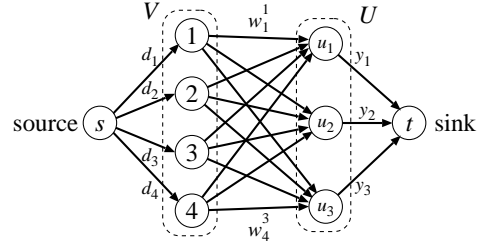

Figure 4: Directed graph $\mathcal{G}_\tau$ associated with a decomposable submodular function $\tau$ with $n=4$ and $k=3$

is equal to

$$\kappa(S) + a(S \cap A_+) + (-a(A_- \setminus S))$$
$$= \kappa(S) + a(S) - a(A_-)$$
$$= \kappa_a(S) + \text{const.}$$

Thus, $\kappa_a$ can be regarded as an $s$-$t$ cut function on $\mathcal{G}_{\boldsymbol{a}}$.

### 3.3 Decomposable submodular functions

Decomposable submodular function (see [31]) are one of the most important special case of generalized graph cut functions [14]. For more examples of generalized graph cut functions, refer to Jegelka et al. [14].

A *decomposable submodular function* $\tau : 2^V \to \mathbb{R}$ is a set function that can be represented as a sum of a modular set function and submodular set functions arising from concave functions. As to Stobbe and Krause [31], we will focus on the case where each concave function is a threshold potential. That is, we consider the following decomposable submodular function $\tau : 2^V \to \mathbb{R}$ defined by

$$\tau(S) = -d(S) + \sum_{j=1}^{k} \min\{y_j, w^j(S)\} \qquad (5)$$

for each $S \subseteq V$, where $\boldsymbol{d} \in \mathbb{R}^n$ is a positive vector, $\boldsymbol{w}^1, \ldots, \boldsymbol{w}^k \in \mathbb{R}^n$ are nonnegative vectors, and $y_1, \ldots, y_k > 0$.

Now we observe that the function $\tau$ defined in (5) can be represented as a generalized graph cut function defined in Subsection 3.1. Consider a directed graph $\mathcal{G}_\tau = (\{s\} \cup \{t\} \cup V \cup U, \mathcal{E})$, where $V = \{1, \ldots, n\}$, $U = \{u_1, \ldots, u_k\}$, and $\mathcal{E} = \{(s,i) : i \in V\} \cup \{(i, u_j) : i \in V, u_j \in U\} \cup \{(u_j, t) : u_j \in U\}$. The edge capacities are determined as

$$\begin{aligned} c(s,i) &= d_i, \quad \forall i \in V, \\ c(i,u_j) &= w^j_i, \quad \forall i \in V, \forall u_j \in U, \\ c(u_j,t) &= y_j, \quad \forall u_j \in U. \end{aligned}$$

Figure 4 illustrates the directed graph $\mathcal{G}_\tau$. We can observe that $\mathcal{G}_\tau$ generates the decomposable submodular function $\tau$.

The function $\tau$ corresponds to the sum of truncated functions described in [14], and the construction of $\mathcal{G}_\tau$ is widely used in computer vision [15].

## 4 Applications

It is known that the convex optimization problem (2) under submodular constraints is related to some discrete and continuous optimization problems. In this section, we show some examples in which the submodular functions have graph structures considered in Section 3.

### 4.1 Finding dense subgraphs

Let $\overline{G} = (V, \overline{E})$ be an undirected graph with node set $V = \{1, \ldots, n\}$ and undirected edge set $\overline{E}$. Given nonnegative edge capacities $c(e)$ ($e \in \overline{E}$) and an integer $k$, the densest $k$-subgraph problem asks for finding a $k$-subset $S \subseteq V$ that maximizes $\theta(S)$, where $\theta(S)$ is the sum of weights of edges in the subgraph induced by $S$. The function $\theta : 2^V \to \mathbb{R}$ is a supermodular function with $\theta(\varnothing) = 0$, and the minimum norm base problem

$$\min_{\boldsymbol{x} \in \mathrm{B}(-\boldsymbol{\theta})} \sum_{i=1}^{n} x_i^2 \qquad (6)$$

plays an important role to find dense subgraphs of $\overline{G}$ [23].

We show that $-\theta$ is a transformed graph cut function (Subsection 3.2). Let $\boldsymbol{m} \in \mathbb{R}^n$ be a vector defined by $m_i = \sum_{i'}\{c(\{i,i'\}) : \{i,i'\} \in \overline{E}\}$ for each $i \in V$, and let $\overline{\kappa}$ be a cut function of $\overline{G}$, that is, $\overline{\kappa}(S) = \sum\{c(\{i,i'\}) : \{i,i'\} \in \overline{E}, i \text{ in } S \text{ and } i' \text{ in } V \setminus S\}$ ($S \subseteq V$). Then we have

$$-\theta(S) = \tfrac{1}{2}\overline{\kappa}(S) - \tfrac{1}{2}m(S)$$

for each $S \subseteq V$. It is easy to see that the function $\overline{\kappa} : 2^V \to \mathbb{R}$ can be regarded as a cut function of a directed graph. Thus, $-\theta$ is a transformed graph cut function.

### 4.2 Proximal methods

Regularized learning is a fundamental formulation for many supervised problems. Let $\{(\boldsymbol{z}_i, y_i)\}_{i=1}^{N}$ be a set of samples, $\boldsymbol{\beta} \in \mathbb{R}^n$ a model parameter vector and $l(\boldsymbol{z}, y; \boldsymbol{\beta})$ a (differentiable) convex loss. Then, the optimization for regularized learning is represented as

$$\min_{\boldsymbol{\beta} \in \mathbb{R}^n} \sum_{i=1}^{N} l(\boldsymbol{z}_i, y_i; \boldsymbol{\beta}) + \lambda \cdot \Omega(\boldsymbol{\beta}),$$

where $\Omega(\boldsymbol{\beta})$ is a regularization term and $\lambda$ is the regularization parameter. If $\Omega(\boldsymbol{\beta})$ is non-differentiable on $\boldsymbol{\beta}$, which is usually true for structured regularization, the proximal method is a popular approach to solve this optimization problem [3]. As is well known, its update procedure at each iteration can be reduced to the calculation of the following problem:

$$\min_{\boldsymbol{\beta} \in \mathbb{R}^n} \frac{1}{2} \|\boldsymbol{\beta} - \boldsymbol{s}\|_2^2 + \lambda \cdot \Omega(\boldsymbol{\beta}), \tag{7}$$

where $\boldsymbol{s} \in \mathbb{R}^n$. Recently, Bach [2] showed that many of the popular structured norms can be represented as continuous relaxations, called Lovász extensions, of submodular functions. And in this case, Problem (7) can be transformed to

$$\min_{\boldsymbol{t}} \{\sum_{i=1}^n t_i^2 : \boldsymbol{t} \in \mathrm{B}(g - \lambda^{-1}s)\},$$

where $g$ is a submodular function whose Lovász extension is $\Omega$ ($\boldsymbol{\beta}$ is solved as $\lambda \boldsymbol{t}$). Note that many of popular structured norms can be expressed as the Lovász extensions of generalized graph cut functions, such as cut functions (that correspond to fused-regularization)[4] and coverage functions (that correspond to overlapping group-regularization).

### 4.3 Minimum ratio problems

For a nonnegative submodular function $g : 2^V \to \mathbb{R}$ with $g(\varnothing) = 0$ and a positive vector $\boldsymbol{b} \in \mathbb{R}^n$, consider the minimum ratio problem which asks for a subset $S \in 2^V \setminus \{\varnothing\}$ minimizing $g(S)/b(S)$. This kind of optimization problems have to be solved iteratively, e.g., in the primal-dual approximation algorithm for a submodular cost covering problem [13].

Suppose that we have the optimal solution $\boldsymbol{x}^*$ to $\min\{\sum_{i=1}^n \frac{x_i^2}{b_i} : \boldsymbol{x} \in \mathrm{B}(f)\}$. Let $\xi_1 = \min_{i \in V} \frac{x_i^*}{b_i}$ and let $S_1 = \{i \in V : \frac{x_i^*}{b_i} = \xi_1\}$. Then the subset $S_1$ is an optimal solution to the minimum ratio problem (see [9]). Therefore, by solving the separable quadratic minimization problem over $\mathrm{B}(g)$, an optimal solution to the minimum ratio problem can be obtained. If the function $g$ has a graph structure, the running time of the approximation algorithm of [13] could be improved.

## 5 A general framework for separable convex minimization under submodular constraints

In this section, we describe the decomposition algorithm, which is a general framework to solve the separable convex minimization problem under submodular constraints. Before describing the decomposition algorithm, we give a parametric formulation of problem (1.a).

For the validity of the decomposition algorithm described here, e.g., refer to Fujishige [9], and Nagano and Aihara [22].

### 5.1 A parametric formulation

Let $f : 2^V \to \mathbb{R}$ be a general nondecreasing submodular function and $\boldsymbol{b} \in \mathbb{R}^n$ a positive vector. Recall that the set function $b$ associated to $\boldsymbol{b}$ is modular.

For a parameter $\alpha \geq 0$, define $f_\alpha : 2^V \to \mathbb{R}$ by $f_\alpha = f - \alpha b$, which is submodular. Let us see how problem (1.a) can be reduced to the parametric submodular minimization problem: minimize $f_\alpha$ for all $\alpha \geq 0$. It is known that there exist $\ell + 1$ subsets,

$$(\varnothing =) S_0 \subset S_1 \subset \cdots \subset S_\ell (= V),$$

and $\ell + 1$ subintervals of $\mathbb{R}_{\geq 0} = \{\alpha \in \mathbb{R} : \alpha \geq 0\}$,

$$R_0 = [0, \alpha_1), \ R_1 = [\alpha_1, \alpha_2), \ \ldots,$$
$$R_j = [\alpha_j, \alpha_{j+1}), \ \ldots, \ R_\ell = [\alpha_\ell, +\infty),$$

such that, for each $j \in \{0, \ldots, \ell\}$, the subset $S_j$ is the unique maximal minimizer of $f_\alpha = f - \alpha b$ for all $\alpha \in R_j$. The vector $\boldsymbol{x}^* \in \mathbb{R}^n$ determined by, for each $i \in V$ with $i \in S_{j+1} \setminus S_j$ ($j \in \{1, \ldots, \ell\}$),

$$x_i^* = \frac{f(S_{j+1}) - f(S_j)}{b(S_{j+1} \setminus S_j)} b_i \tag{8}$$

is the unique optimal solution to the quadratic minimization problem (1.a). The equation (8) implies that problem (1.a) can be immediately solved if the collection $\mathcal{S}^* = \{S_0, S_1, \ldots, S_\ell\}$ is computed.

### 5.2 The decomposition algorithm

By successively minimizing $f_\alpha = f - \alpha b$ for some appropriately chosen $\alpha \geq 0$, the decomposition algorithm finds $S_j$ one by one, and finally the chain $S_0 \subset S_1 \subset \cdots \subset S_\ell$ and the optimal solution $\boldsymbol{x}^*$ to problem (1.a) are obtained.

The decomposition algorithm DA is recursive. Suppose that we are given two subsets $S_j, S_{j'} \in \mathcal{S}^*$ with $0 \leq j < j' \leq n$. The algorithm $\mathsf{DA}(S_j, S_{j'})$ finds the collection

$$\mathcal{S}^*(S_j, S_{j'}) := \{S \in \mathcal{S}^* : S_j \subseteq S \subseteq S_{j'}\}.$$

It can be verified that $\alpha_{j+1} \leq \frac{f(S_{j'})-f(S_j)}{b(S_{j'}\setminus S_j)} \leq \alpha_{j'}$. Therefore, we can decide if $(j+1 = j')$ or $(j+1 < j')$ by minimizing $f_\alpha$ with $\alpha = \frac{f(S_{j'})-f(S_j)}{b(S_{j'}\setminus S_j)}$.

The decomposition algorithm DA can be described as follows (see, e.g., [22] for the detailed analysis of the algorithm).

---

**Algorithm** DA$(T, T')$
*Input:* Subsets $T, T' \in \mathcal{S}^*$ with $T \subset T'$.
*Output:* The collection $\mathcal{S}^*(T, T')$.

1: Set $\alpha = \frac{f(T')-f(T)}{b(T'\setminus T)}$. Compute the unique maximal minimizer $T''$ of $f_\alpha := f - \alpha b$.
2: If $T'' = T'$, return $\{T, T'\}$.
3: If $T \subset T'' \subset T'$, let $\mathcal{S}_1$ and $\mathcal{S}_2$ be the collections returned by DA$(T,T'')$ and DA$(T'',T')$, respectively. Return $\mathcal{S}_1 \cup \mathcal{S}_2$.

---

First of all, we know that $S_0 = \varnothing$ and $S_\ell = V$, although we do not know how large $\ell$ is. Clearly, we have $\mathcal{S}^*(\varnothing, V) = \mathcal{S}^*$. Therefore, by performing DA$(\varnothing, V)$, the collection $\mathcal{S}^*$ can be obtained. Using (8), we can immediately obtain the optimal solution of problem (1.a).

In the decomposition algorithm DA$(\varnothing, V)$, we minimize the functions $f_\alpha: 2^V \to \mathbb{R}$ at most $2n-1$ times.

## 6  Efficient algorithms for structured convex minimization problems

Let $\gamma: 2^V \to \mathbb{R}$ be a generalized graph cut function defined as in (4), which is generated from a directed graph $\mathcal{G} = (\{s\}\cup\{t\}\cup V\cup U, \mathcal{E})$. Consider the convex optimization problem (1.a) with $f = \gamma$,

$$\min_{\boldsymbol{x}\in B(\gamma)} \sum_{i=1}^n \frac{x_i^2}{b_i}. \quad (9)$$

Recall that $\boldsymbol{b} \in \mathbb{R}^n$ is a positive vector. We show that problem (9) can be solved efficiently using the framework of the decomposition algorithm DA of Section 5.

For nonnegative parameters $\alpha$ and $\beta$, let us see that the set functions $\gamma - \alpha b$ and $\gamma + \beta b$ are both generalized graph cut functions. By adding new edges $e_i^- = (s, i)$ ($i \in V$) with edge capacities $c(e_i^-) = \alpha b_i$ ($i \in V$) to $\mathcal{G}$, we construct a new directed graph $\mathcal{G}_{\alpha\boldsymbol{b}}^-$ (see Figure 5 (a)). By adding new edges $e_i^+ = (i, t)$ ($i \in V$) with edge capacities $c(e_i^+) = \beta b_i$ ($i \in V$) to $\mathcal{G}$, we construct a new directed graph $\mathcal{G}_{\beta\boldsymbol{b}}^+$ (see Figure 5 (b)). Since $\gamma$ is defined as in (4), the functions $\gamma - \alpha b$ and $\gamma + \beta b$ are generated by $\mathcal{G}_{\alpha\boldsymbol{b}}^-$ and $\mathcal{G}_{\beta\boldsymbol{b}}^+$, respectively.

Using Lemmas 2 and 3, and the fact that $\gamma + \beta b$ is a generalized graph cut function, we can assume that $\gamma$ is nondecreasing in problem (9).

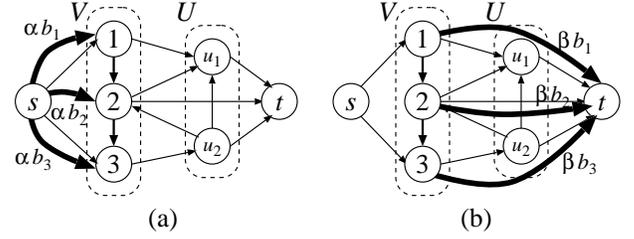

Figure 5: Directed graphs $\mathcal{G}_{\alpha\boldsymbol{b}}^-$ and $\mathcal{G}_{\beta\boldsymbol{b}}^+$

Now we can apply the decomposition algorithm DA$(\varnothing, V)$ to problem (9). In step 1 of the algorithm DA, we just have to compute the maximal minimum $s$-$t$ cut in $\mathcal{G}_{\alpha\boldsymbol{b}}^-$ for some appropriately chosen $\alpha \geq 0$ to find the maximal minimizer of $\gamma - \alpha b$. Since we minimize the functions $\gamma - \alpha b$ at most $2n-1$ times, we obtain the following theorem with the aid of the minimum $s$-$t$ cut algorithm [28].

**Theorem 6.** *For a generalized graph cut function $\gamma: 2^V \to \mathbb{R}$ generated from $\mathcal{G} = (\{s\}\cup\{t\}\cup V\cup U, \mathcal{E})$, problem (9) can be solved in $\mathrm{O}(n(n+|U|)|\mathcal{E}|)$ time, where $n = |V|$.*

We can obtain a different time complexity by using the parametric minimum cut algorithm (Gallo et al. [11]). The parametric minimization problem

$$\text{minimize } \gamma - \alpha b \text{ for all } \alpha \geq 0$$

corresponds to the parametric minimum cut problem

$$\text{find minimum } s\text{-}t \text{ cuts in } \mathcal{G}_\alpha^- \text{ for all } \alpha \geq 0.$$

To solve this parametric cut problem, we can utilize the parametric minimum cut algorithm [11] (see also [16]). We remark that the directed graph $\mathcal{G}_\alpha^-$ satisfies the monotonicity in $\alpha \geq 0$ in the meaning of [11]. As a result, we have the following time complexity.

**Theorem 7.** *For a generalized graph cut function $\gamma: 2^V \to \mathbb{R}$ generated from $\mathcal{G} = (\{s\}\cup\{t\}\cup V\cup U, \mathcal{E})$, problem (9) can be solved in $\mathrm{O}((n+|U|)|\mathcal{E}|\log\frac{(n+|U|)^2}{|\mathcal{E}|})$ time, where $n = |V|$.*

The algorithm of Theorem 7, which is much faster than that of Theorem 6 from a theoretical point of view, is rather complicated to implement.

In view of Theorem 1, we can solve the convex minimization problem under constraints with respect to the structured submodular function $\gamma$,

$$\min_{\boldsymbol{x}\in B(\gamma)} \sum_{i\in V} w_i(x_i)$$

in $O(n(n+|U|)|\mathcal{E}|)$ or $O((n+|U|)|\mathcal{E}|\log\frac{(n+|U|)^2}{|\mathcal{E}|})$ time for a number of separable convex objective functions.

# 7 Experimental results

We investigated the empirical performance of the proposed scheme using synthetic and real-world datasets. In Section 7.1, we compare the proposed method in the application to proximal methods for structured regularized least-squares regression, with the state-of-the-art algorithms. In Section 7.2, we apply the proposed algorithm to the densest subgraph problem for large real web-network data. The experiments below were run on a 2.3 GHz 64-bit workstation using Matlab with Mex implementations. And we used SPAMS (SPArse Modeling Software) [18] for the implementations of the proximal methods for the first experiment.

## 7.1 Comparison in proximal methods

In the first experiment, we compared the proposed algorithm in the application to proximal methods with the state-of-the-art algorithms. As for the regularization term, we used fused-regularization $\Omega_{\text{fused}}(\boldsymbol{\beta})$ and group regularization (with $l_\infty$-norm) $\Omega_{\text{group}}(\boldsymbol{\beta})$ (for a given set of groups $\mathcal{G}$), respectively represented as

$$\Omega_{\text{fused}}(\boldsymbol{\beta}) = \sum_{i=1}^{n-1}|\beta_i - \beta_{i+1}| \text{ and}$$
$$\Omega_{\text{group}}(\boldsymbol{\beta}) = \sum_{g\in\mathcal{G}}d_g\|\boldsymbol{\beta}_g\|_\infty,$$

where $d_g$ is the weight of the group $g$. As the comparison partners, we used the proximal methods for the above regularization; the one based on the homotopy algorithm for $\Omega_{\text{fused}}(\boldsymbol{\beta})$ [7] (Homo.) and the one by Mairal et al. [19] for $\Omega_{\text{group}}(\boldsymbol{\beta})$ (NFA), as well as the minimum-norm-point algorithm (MNP) for the calculation of the proximal operator. Since both regularizations can be represented as the decomposable submodular function, we applied the parametric flow algorithm for computing the proximal operators (DA).

We generated data as follows. First for the evaluation with fused regularization, one feature is first selected randomly and the next one is selected with probability 0.4 from each neighboring feature or with probability $0.2/(N-2)$ from the remaining ones and repeat this procedure until $k$ features are selected. For group regularization, the features are covered by 20–200 overlapping groups of size 15. The causal features are chosen to be the union of 2 of these groups. Here, we assign weights $d_g = 2$ to those causal groups and $d_g = 1$ to all other groups. We then simulate $N$ data points $(\mathbf{x}(i), y(i))$, with $y(i) = \bar{\boldsymbol{\beta}}^\top \mathbf{x}(i) + \epsilon$, $\epsilon \sim \mathcal{N}(0, \sigma^2)$, where $\bar{\boldsymbol{\beta}}$ is 0 for non-causal features and normally distributed otherwise.

Table 1: Comparison of running time (seconds) for the proposed and existing methods.

| $n$ | $N$ | $k$ | DA | MNP | Homo. |
|---|---|---|---|---|---|
| 500 | 500 | 20 | 0.024 | 5.083 | 0.084 |
| 500 | 1,000 | 20 | 0.062 | 146.969 | 0.531 |
| 500 | 5,000 | 20 | 1.085 | — | 32.676 |
| 1,000 | 500 | 20 | 0.019 | 3.891 | 0.058 |
| 1,000 | 1,000 | 20 | 0.059 | 98.310 | 0.266 |
| 1,000 | 5,000 | 20 | 1.064 | — | 12.372 |

| $n$ | $N$ | $k$ | DA | MNP | NFA |
|---|---|---|---|---|---|
| 500 | 500 | ∼20 | 0.021 | 8.910 | 0.015 |
| 500 | 1,000 | ∼20 | 0.056 | 280.117 | 0.052 |
| 500 | 5,000 | ∼20 | 1.091 | — | 1.112 |
| 1,000 | 500 | ∼20 | 0.020 | 6.108 | 0.015 |
| 1,000 | 1,000 | ∼20 | 0.054 | 198.010 | 0.051 |
| 1,000 | 5,000 | ∼20 | 1.003 | — | 0.896 |

Since all methods calculate the same objectives in principle, here we report only the comparison of the empirical running time. Tables 1 show the running time by the algorithms for reaching the duality-gap within $10^{-4}$, averaged over 20 datasets each. We can see that the algorithms based on the parametric-flow algorithm, including ours, run much faster than the others. Note that our scheme can be applied to more general form of structured regularization (Eq. (5)) for the graph cut implementation) than $\Omega_{\text{fused}}(\boldsymbol{\beta})$ and $\Omega_{\text{group}}(\boldsymbol{\beta})$.

## 7.2 Densest subgraphs in web graphs

In the second experiment, we applied the proposed algorithm to the densest subgraph problem using public web-graph and social-network datasets [5]. The characteristics of each data set are shown in Table 2. Although the minimum-norm-point algorithm was applied to the same problem on one of the datasets (*cnr-2000*) in [23], the data was sub-sampled to 5,000 nodes due to its computational cost. However, in this experiment, we used the full datasets for the analyses, which was possible because our framework runs much more efficiently than the algorithm in [23].

The running time for applying our method to each dataset is shown in Table 2 as well as the number of optimal solutions found by the algorithm. Our method could find exactly optimal-solutions for several $k$ for these large datasets in practical time. Note again that, if $k$ is fixed beforehand, the densest subgraph problem with the size constraint is NP-hard and thus there is no efficient algorithm. Also, the graphs in Figure 6 show plot examples of intensity $I(\mathcal{S})$ versus the sizes of subsets $k$ found by the algorithm. The tendency seems to be that our methods can find more optimal solutions if graphs are denser.

Table 2: Resulting running-time and the number of optimal subsets found by the algorithm as well as the characteristics of datasets.

| Data | # Node | # Arc | Time [s] | # Set |
|---|---|---|---|---|
| cnr-2000 | 325,557 | 3,216,152 | 20.55 | 22 |
| uk-2007 | 100,000 | 3,050,615 | 19.70 | 49 |
| in-2004 | 1,382,908 | 16,917,053 | 225.90 | 5,971 |
| eu-2500 | 862,664 | 19,235,140 | 278.50 | 4,933 |
| wordassoc. | 10,617 | 72,172 | 0.15 | 2 |
| amazon-2008 | 735,323 | 5,158,388 | 127.51 | 1,882 |
| dblp-2010 | 326,186 | 1,615,400 | 19.68 | 985 |
| dblp-2011 | 986,324 | 6,707,236 | 96.60 | 979 |

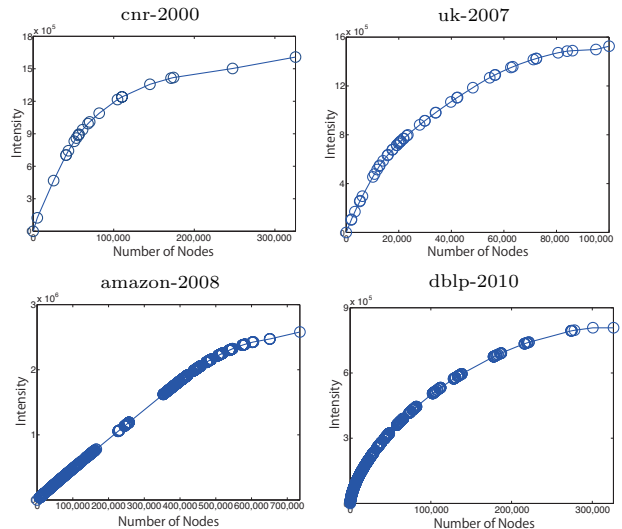

Figure 6: Example plots of $I(\mathcal{S})$ versus $k$ for *cnr-2000*, *uk-2007*, *amazon-2008* and *dblp-2010*.

## 8   Concluding remarks

We have shown that when a submodular function $f$ has a directed graph representation the separable convex minimization problem under submodular constraints can be solved pretty efficiently compared to general submodular optimization methods. It is known that quite a lot of submodular functions have graph structures (refer to Jegelka, Lin, and Bilmes [14]). The proposed methods are based on the general theory of submodular functions and (parametric) maximum flow algorithms. In addition, we remark that the proposed methods can deal with various essentially equivalent objective functions for the problem.

**Appendix: Submodularity of generalized graph cut functions**

In order to make this paper self-contained, we give a proof of the submodularity of a generalized graph cut function [14], $\gamma : 2^V \to \mathbb{R}$ defined in (4).

We set $\beta \geq 0$ as the sum of all edge capacities of $\mathcal{G}$. Let $\mathbf{1} \in \mathbb{R}^n$ be the all-one vector and let $1 : 2^V \to \mathbb{R}$ be a set function defined by $1(S) = |S|$ for each $S \subseteq V$. The directed graph $\mathcal{G}^+_{\beta\mathbf{1}}$ (see Section 6) generates the set function $\gamma + \beta 1$. For each $S \subseteq V$, let $\gamma'(S)$ be the minimum capacity of a cut separating $\{s\} \cup S$ from the sink $t$ in $\mathcal{G}^+_{\beta\mathbf{1}}$. By the result of Megiddo [20] on network flow problems with multiple terminals, the set function $\gamma' : 2^V \to \mathbb{R}$ is submodular. For each $S \subseteq V$, we have

$$\gamma'(S) = \min_{W \subseteq (U \cup V \setminus S)} \sum \{c(e) : e \in \delta^{\text{out}}_{\mathcal{G}^+_{\beta\mathbf{1}}}(\{s\} \cup S \cup W)\}$$

$$= \min_{W \subseteq (U \cup V \setminus S)} \Big( \sum \{c(e) : e \in \delta^{\text{out}}_{\mathcal{G}}(\{s\} \cup S \cup W)\}$$

$$+ \beta|S| + \beta|W \cap (V \setminus S)|\Big)$$

$$= \beta|S| + \min_{W \subseteq U} \sum \{c(e) : e \in \delta^{\text{out}}_{\mathcal{G}}(\{s\} \cup S \cup W)\}$$

$$= \beta|S| + \gamma(S),$$

where the third equality holds because $\beta$ is sufficiently large. Since $\gamma' = \gamma + \beta 1$ is submodular, the function $\gamma$ is also submodular.

**Acknowledgments**

This research was partially supported by Aihara Project, the FIRST program from JSPS, JST ERATO Kawarabayashi Large Graph Project, JST PRESTO PROGRAM (Synthesis of Knowledge for Information Oriented Society), and the Cooperative Research Program of "Network Joint Research Center for Materials and Devices".